\documentclass[lettersize,journal]{IEEEtran}
\usepackage{amsmath,amsfonts}
\usepackage{algorithmic}
\usepackage[ruled,vlined]{algorithm2e}
\usepackage{array}
\usepackage[caption=false,font=normalsize,labelfont=sf,textfont=sf]{subfig}
\usepackage{textcomp}
\usepackage{stfloats}
\usepackage{url}
\usepackage{verbatim}
\usepackage{graphicx}
\usepackage{cite}
\usepackage{tikz}
\usepackage{jabbrv}
\usepackage{amssymb}   
\usepackage{pifont}    
\usepackage{booktabs}
\usepackage{multirow}
\usepackage{xcolor}
\usepackage{tabularx}
\usepackage{makecell}
\usepackage{float}
\usepackage{placeins}
\usepackage{ragged2e}
\usepackage{subcaption}
\usepackage{multirow}
\usepackage{booktabs}
\usepackage{makecell}   
\usepackage{amssymb}
\usepackage{pifont}

\newcolumntype{?}{!{\vrule width 1pt}}

\begin{document}

\title{Sparse Threats, Focused Defense: Criticality-Aware Robust Reinforcement Learning for Safe Autonomous Driving
}

\author{Qi Wei, Junchao Fan, Zhao Yang, Jianhua Wang, Jingkai Mao, and Xiaolin Chang
\thanks{Qi Wei, Junchao Fan, Zhao Yang, Jingkai Mao, and Xiaolin Chang are with the Beijing Key Laboratory of Security and Privacy in Intelligent Transportation, Beijing Jiaotong University, P.R.China. (e-mail: \{25121689, 23111144, 24119018, 23111143, xlchang\}@bjtu.edu.cn)

Jianhua Wang is with the College of Computer Science and Technology, Taiyuan University of Technology, P.R.China. (e-mail: wangjianhua02@tyut.edu.cn)
}
\thanks{Manuscript received April 19, 2021; revised August 16, 2021.}}

\markboth{Journal of \LaTeX\ Class Files,~Vol.~14, No.~8, August~2021}%
{Shell \MakeLowercase{\textit{et al.}}: A Sample Article Using IEEEtran.cls for IEEE Journals}

\IEEEpubid{0000--0000/00\$00.00~\copyright~2021 IEEE}

\maketitle

\IEEEpubid{}

\begin{abstract}
\justifying
Reinforcement learning (RL) has shown considerable potential in autonomous driving (AD), yet its vulnerability to perturbations remains a critical barrier to real-world deployment.
As a primary countermeasure, adversarial training improves policy robustness by training the AD agent in the presence of an adversary that deliberately introduces perturbations.
Existing approaches typically model the interaction as a zero-sum game with continuous attacks. 
However, such designs overlook the inherent asymmetry between the agent and the adversary and then fail to reflect the sparsity of safety-critical risks, rendering the achieved robustness inadequate for practical AD scenarios.
To address these limitations, we introduce criticality-aware robust RL (CARRL), a novel adversarial training approach for handling sparse, safety-critical risks in autonomous driving.
CARRL consists of two interacting components: a risk exposure adversary (REA) and a risk-targeted robust agent (RTRA).
We model the interaction between the REA and RTRA as a general-sum game, allowing the REA to focus on exposing safety-critical failures (e.g., collisions) while the RTRA learns to balance safety with driving efficiency.
The REA employs a decoupled optimization mechanism to better identify and exploit sparse safety-critical moments under a constrained budget.
However, such focused attacks inevitably result in a scarcity of adversarial data.
The RTRA copes with this scarcity by jointly leveraging benign and adversarial experiences via a dual replay buffer and enforces policy consistency under perturbations to stabilize behavior.
Experimental results demonstrate that our approach reduces the collision rate by at least 22.66\% across all cases compared to state-of-the-art baseline methods.

\end{abstract}


\maketitle
\section{Introduction}
Deep reinforcement learning (DRL) has become a core paradigm for autonomous driving (AD), enabling agents to learn complex decision-making policies from high-dimensional observations and optimize long-term objectives in interactive traffic environments~\cite{huang2025vlm, lu2025preference, zhang2024integration, wang2025highway, huang2024human}. 
However, DRL-based driving policies remain highly vulnerable to observation perturbations or perceptual uncertainties, which can induce severe control deviations and lead to safety-critical failures. This vulnerability raises serious concerns regarding the safe and reliable deployment of DRL-based AD policies in real-world environments~\cite{ibrahum2024deep,wang2025, heTrustworthyAutonomousDriving2024a}.

To mitigate these risks, existing research systematically simulates such threats via adversarial attacks, using an artificial adversary to inject perturbations into the agent’s observations to expose vulnerabilities of DRL-based driving policies~\cite{wang2025uncertainty, zhaoSurveyRecentAdvancements2024a, wu2024recent}.
Consequently, adversarial training has emerged as a dominant approach to improve robustness by exposing policies to challenging perturbations during policy learning~\cite{standen2025adversarial, hu2025toward, pelekis2025adversarial, schott2024robust}.
Despite the progress achieved by existing adversarial training methods, several fundamental limitations remain.

First, most prior work formulates the interaction between the agent and the adversary as a zero-sum game~\cite{He2023RobustDM, wang2025, guoRobustTrainingMultiagent2025}. 
However, in AD, the agent’s objective is inherently multi-dimensional, encompassing not only safety but also efficiency, comfort, and task completion. 
In contrast, a realistic adversary should primarily aim to induce safety-critical failures. 
The zero-sum assumption conflates these inherently mismatched objectives and may distort the adversary away from truly worst-case-oriented attack policies.
As a result, the adversary may generate non-critical perturbations, which ultimately diminish the effectiveness of adversarial training.

Second, the majority of existing approaches model adversarial perturbations as continuous perturbations~\cite{wangExplainableDeepAdversarial2024a, Fan2025RobustDC}. 
In real-world driving, however, safety risks are typically concentrated in sparse, critical moments rather than occurring uniformly over time~\cite{yan2023learning}.
Such continuous attack settings are not only unrealistic but also bias the learning process by exposing the agent to persistent perturbations.

While emerging research has successfully demonstrated the efficacy of criticality-aware attack policies that exploit these sparse vulnerabilities~\cite{fan2025less, Fan2025SharpeningTS}, corresponding defense mechanisms remain largely unexplored. 
Focusing adversarial training on safety-critical moments is more effective than applying it uniformly at every timestep, as it concentrates defensive capacity where failures actually occur rather than enforcing unnecessary caution throughout the trajectory.
Nevertheless, this training paradigm introduces new challenges: 
(i) the reduced attack frequency significantly limits the availability of adversarial samples, making it difficult for the agent to leverage them to learn robust policies~\cite{yichao2025enhancing}; and (ii) the coexistence of benign and adversarial experiences requires carefully structured training constraints to ensure stable policy convergence without distorting driving behavior~\cite{liu2023towards}.

To address these gaps, we propose criticality-aware robust reinforcement learning (CARRL), which explicitly focuses robustness on sparse yet safety-critical moments. 
CARRL consists of two interacting components: a risk exposure adversary (REA) and a risk-targeted robust agent (RTRA).
We first model the interaction between the REA and the RTRA as a general-sum Markov game (GMG).
This formulation empowers the REA to focus strictly on identifying and exploiting safety-critical vulnerabilities, while allowing the RTRA to simultaneously optimize for safety and efficiency. 
Regarding the REA, we design it to learn a criticality-aware attack policy that determines \textit{when} and \textit{how} to attack. 
Since this focused attack strategy inherently results in sparse adversarial data, the RTRA is correspondingly engineered to overcome this scarcity, ensuring reliable driving behavior under both benign and critical conditions.

\begin{figure}[!htbp]
    \centering
    \includegraphics[width=\linewidth, trim=5 10 10 10, clip]{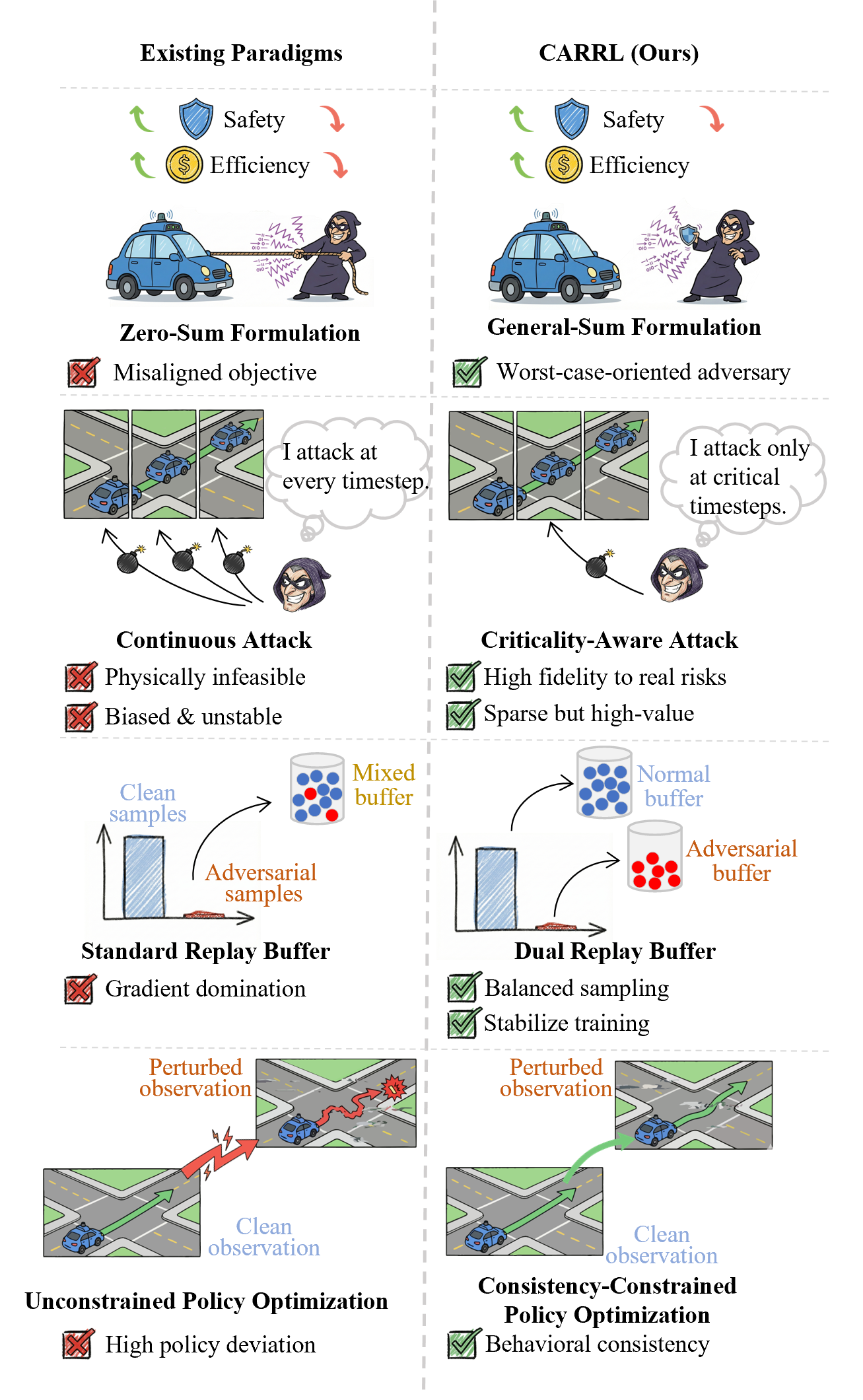}
    \caption{Conceptual comparison highlighting the advantages of CARRL over existing paradigms.}
    \label{fig:framework}
\end{figure}

Fig.~\ref{fig:framework} illustrates CARRL in comparison with existing paradigms and highlights its key innovations.
\textbf{To the best of our knowledge}, CARRL is the first approach to address sparse yet safety-critical risks in AD through focusing defense on critical moments.
CARRL establishes a novel paradigm for robust AD, transforming the defense policy from generalized to focused and demonstrating the potential of criticality-aware robustness in advancing safety in real-world driving.

The main contributions of this paper are summarized as follows:
\begin{itemize}
  \item \textbf{General-Sum Markov Game Formulation}: We propose CARRL, a novel DRL-based approach for robust AD, which formulates the interaction between the RTRA and the REA as a GMG. This formulation overcomes the limitations of zero-sum assumptions, empowering the REA to pursue a strictly worst-case-oriented strategy while enabling the RTRA to balance safety with efficiency.

  \item \textbf{Risk Exposure Adversary with Decoupled Optimization}: We propose the REA designed to exploit safety-critical vulnerabilities within a limited attack budget. 
  To address the conditional dependency in the REA’s action decisions, we introduce a decoupled optimization mechanism that filters out irrelevant gradients at non-attack timesteps, significantly enhancing the precision and efficiency of the REA’s learning. As a result, the REA learns a criticality-aware attack policy that selectively targets the most safety-critical moments.

  \item \textbf{Risk-Targeted Agent with Dual Replay Buffer and Consistency-Constrained Policy Optimization}: To address the scarcity of adversarial data caused by the design of the REA, we introduce the RTRA equipped with two key mechanisms: a dual replay buffer (DRB) and consistency-constrained policy optimization (CCPO). The DRB rebalances the distribution of benign and adversarial experiences to mitigate data imbalance, while CCPO ensures behavioral consistency under adversarial perturbations. This joint design effectively ensures stable policy convergence and robust driving performance.
\end{itemize}

Extensive empirical evaluations under varying perturbation magnitudes, traffic densities, and attack methods validate the superior robustness and strong generalization capabilities of CARRL.
Under default traffic density and perturbation magnitude, our approach achieves a collision rate of 3.67\%, representing a 46.1\% reduction compared to baselines.
Across varying traffic densities, CARRL consistently maintains a lead of at least 26.5\% in success rate.
Notably, our approach exhibits robustness not only against the proposed attacks but also generalizes well to continuous attack settings.
In such scenarios, it matches the success rate of state-of-the-art methods while achieving a 26.83\% reduction in collision rate, underscoring its superior robust performance.

The remainder of this paper is organized as follows. Section~\ref{sec:related} reviews related work. Section~\ref{sec:system} presents the system model, followed by the implementation of CARRL in Section~\ref{sec:method}. Section~\ref{sec:experiments} provides experimental results, and Section~\ref{sec:conclusion} concludes the paper.

\section{Related Work}
\label{sec:related}

Recent advances in DRL have shown strong potential for AD~\cite{wang2024research,zhaoSurveyRecentAdvancements2024a,hoel2019combining}, yet learned policies remain vulnerable to adversarial perturbations and safety-critical uncertainties. 
Consequently, prior work has explored adversarial attack methods to reveal policy weaknesses and robust learning strategies to improve safety. 
This section reviews representative studies along these two lines in Sections~\ref{sec:related1} and~\ref{sec:related2}.

\subsection{Adversarial Attacks for DRL-based AD}
\label{sec:related1}
Recent research has extensively demonstrated the vulnerability of DRL policy in AD to adversarial perturbations~\cite{pan2019characterizing,buddareddygari2022targeted, Fan2025SharpeningTS}. By characterizing the susceptibility of driving agents to engineered perturbations in the observation space or environmental dynamics, these studies reveal that even well-trained policies can be misled into safety-critical failures, such as deviating from lanes or colliding with obstacles.
For instance, Buddareddygari~\emph{et al.}~\cite{buddareddygari2022targeted} exploit differentiable environment models to optimize a static physical visual patch, effectively steering the victim vehicle toward a target state. Similarly, Pan~\emph{et al.}~\cite{pan2019characterizing} investigate both finite-difference based observation perturbations and an RL-based dynamics attacker that manipulates road friction and bump profiles to destabilize the agent. 
While these approaches expose DRL vulnerabilities, they often rely on impractical continuous interference. 
Since safety-critical situations in AD occur infrequently, focusing attacks on these critical moments is more effective than applying indiscriminate perturbations. As a result, recent studies have shifted toward sparse attack policies.
For instance, Fan~\emph{et al.}~\cite{Fan2025SharpeningTS} proposed an adaptive framework for sparse attacks under strict budget constraints. By leveraging an expert-guided reinforcement learning mechanism, their method exposes policy vulnerabilities in complex scenarios using only a small number of attacks, achieving effectiveness comparable to continuous attack policies. Such results suggest that safety failures in AD are often dominated by a small number of critical states or timing windows, highlighting the importance of identifying these moments.

While the above studies effectively demonstrate the vulnerability of DRL-based driving policies, they exhibit clear limitations in terms of defense modeling and practical realism. In particular, defense mechanisms against criticality-aware attacks~\cite{fan2025less} remain largely underexplored. As a result, learning robust and reliable driving policies that can withstand sparse, well-timed adversarial disturbances remains an open challenge.

\subsection{Robust DRL-based AD Methods Against Adversarial Attacks}
\label{sec:related2}
Adversarial training has emerged as a predominant paradigm for enhancing the robustness of DRL-based AD policies. In this line of research, the defense problem is typically formulated as a zero-sum game, where the agent seeks to maximize its reward while an adversary aims to minimize it. A representative example is robust adversarial reinforcement learning~\cite{pinto2017robust}, which models perturbations as an opposing player and trains the agent under a minimax objective. Building on this paradigm, subsequent studies further adopt zero-sum Markov game formulations to model perturbations, resulting in conservative yet robust driving policies~\cite{he2022robust,wang2025}. For instance, He~\emph{et al.}~\cite{he2022robust} formulate AD as a zero-sum Markov game between the ego vehicle and an adversarial environment. In their approach, environmental perturbations are modeled as adversarial actions that directly oppose the agent’s objective, and the driving policy is optimized against worst-case perturbations. This zero-sum formulation enables robust performance guarantees under adversarial conditions.

Despite their effectiveness, zero-sum assumptions can be overly restrictive for AD.
In practice, an adversary is typically focused on inducing safety-critical failures, whereas the agent must balance efficiency and safety.
This has motivated a shift toward general-sum modeling, in which the agent and the adversary pursue distinct objectives~\cite{heTrustworthyAutonomousDriving2024a}. 
For instance, Fan~\emph{et al.}~\cite{Fan2025RobustDC} explicitly model asymmetric goals and constraints between the ego vehicle and the adversary to improve robust driving control.

However, the aforementioned defense methods are primarily designed to counter unconstrained, continuous adversarial attacks, which rarely reflect real-world AD scenarios.
In practice, perturbations result in catastrophic failures primarily when they occur at a small number of critical moments.
Consequently, prioritizing robustness at these critical moments yields substantially higher safety improvements than applying uniform defense across all timesteps.
While emerging research has begun to explore criticality-aware attack policies that exploit these vulnerabilities, the development of corresponding defense mechanisms remains largely unexplored.

Motivated by these gaps, we propose CARRL, a new adversarial training approach that introduces an REA capable of executing criticality-aware attacks specifically designed to induce safety-critical failures.
To address the resulting scarcity of adversarial samples, CARRL equips the RTRA with a DRB and CCPO, enabling stable policy learning and robust driving performance.
\section{System Model}
\label{sec:system}
\begin{figure}[t]
  \centering
  \includegraphics[width=1.0\linewidth]{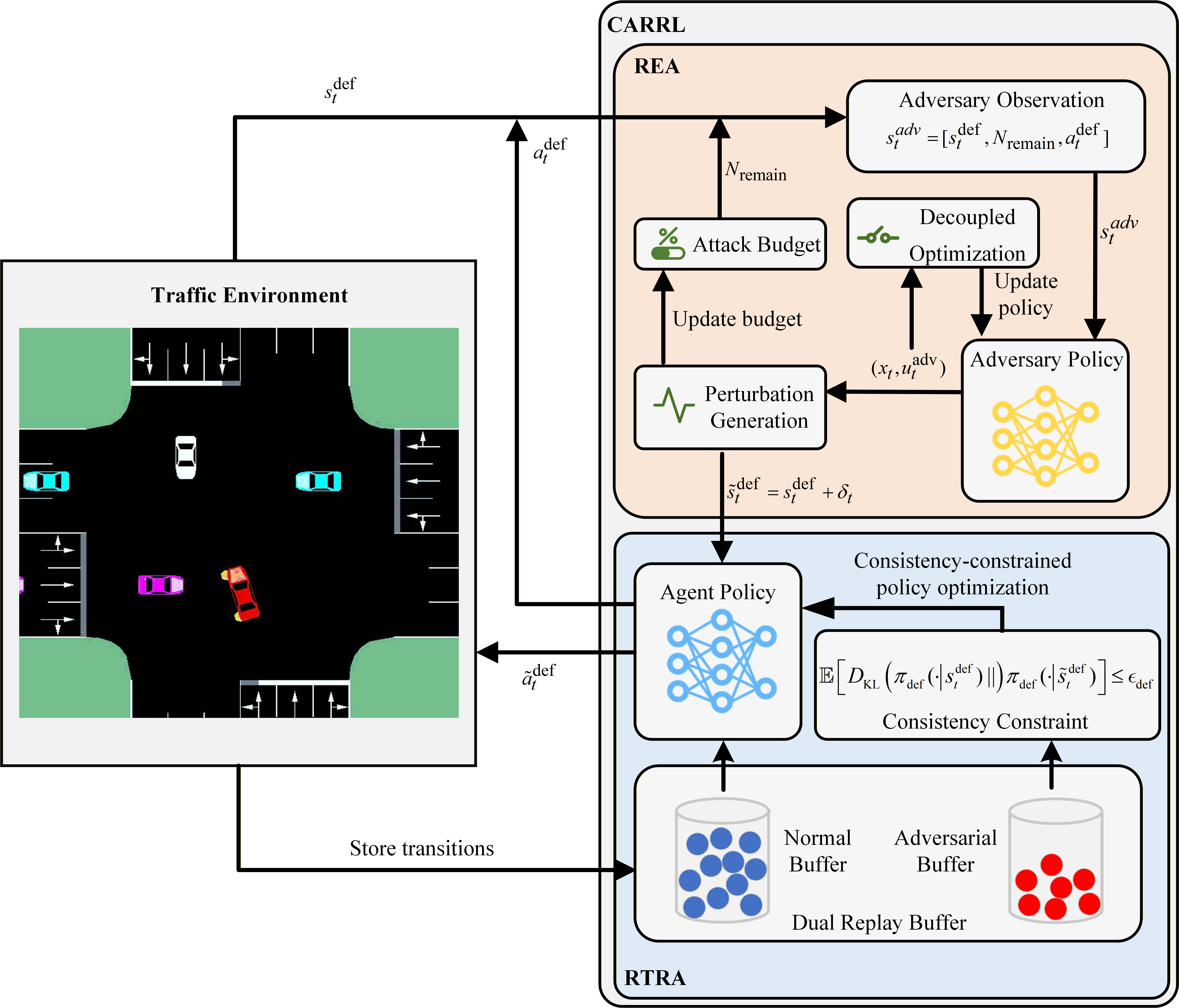}
  \caption{The system model of CARRL, which consists of two components, namely, REA and RTRA. Under a limited attack budget, the REA learns to trigger perturbations at safety-critical moments, while the RTRA learns a robust driving policy by jointly leveraging benign and adversarial experiences.}
  \label{fig:system}
\end{figure}

This section formally characterizes the problem of robust AD subject to rare but structured perturbations. 
As illustrated in Fig.~\ref{fig:system}, we model this problem through the interaction of the two components of REA and RTRA. 
The REA generates perturbations at critical moments to lead the RTRA to safety-critical failures, while the RTRA controls the red ego vehicle to achieve driving objectives and maintain robustness against these perturbations.
Given their asymmetric objectives, the interaction between the REA and the RTRA is formulated as a two-player GMG tailored for DRL.
The formulation of this game is established in Section~\ref{sec:problem}. Building upon this, the specific designs of the REA and the RTRA are detailed in Sections~\ref{sec:adversary_3} and \ref{sec:agent_3}, respectively.

\subsection{Two-player General-Sum Markov Game}
\label{sec:problem}
We formulate the interaction between the REA and the RTRA as a two-player general-sum Markov game.
Formally, it is defined by the tuple
\begin{equation}
(\mathcal{S}^{\mathrm{def}}, \tilde{\mathcal{S}}^{\mathrm{def}}, \mathcal{S}^{\mathrm{adv}}, \mathcal{A}^{\mathrm{def}}, \mathcal{A}^{\mathrm{adv}}, P, r^{\mathrm{def}}, r^{\mathrm{adv}}, \gamma, T).
\end{equation}
$\mathcal{S}^{\mathrm{def}}$ and $\tilde{\mathcal{S}}^{\mathrm{def}}$ denote the clean and perturbed observation spaces of the RTRA, respectively.
$\mathcal{S}^{\mathrm{adv}}$ represents the observation space of the REA.
$\mathcal{A}^{\mathrm{def}}$ and $\mathcal{A}^{\mathrm{adv}}$ are the action spaces of the RTRA and the REA, respectively.
$r^{\mathrm{def}}$ and $r^{\mathrm{adv}}$ denote the reward functions associated with the RTRA and the REA, respectively.
$P$ denotes the state transition probability function, formally defined as $P: \mathcal{S}^{\mathrm{def}} \times \mathcal{A}^{\mathrm{def}} \times \mathcal{A}^{\mathrm{adv}} \times \mathcal{S}^{\mathrm{def}} \to [0, 1]$, which governs the probability $P(s^{\mathrm{def}}_{t+1} \mid s^{\mathrm{def}}_t, a_t^{\mathrm{def}}, a_t^{\mathrm{adv}})$ of the environment evolving to state $s_{t+1}$ given the current state $s_t$ and the joint action pair $(a_t^{\mathrm{def}}, a_t^{\mathrm{adv}})$.
$\gamma \in [0,1]$ is the discount factor and $T$ denotes the finite time horizon.

Sections~\ref{sec:adversary_3} and \ref{sec:agent_3} present problem-specific specifications of the key elements based on the above formulation.

\subsection{Optimization Problem formulation of REA}
\label{sec:adversary_3}
Departing from static or heuristic perturbation patterns, our REA is designed to learn an adaptive attack policy, preventing the RTRA from overfitting to predefined attack patterns.
A finite attack budget is imposed because real-world perturbations are infrequent, and continuous perturbations would otherwise dominate the learning process of the RTRA, leading to unstable AD policy learning.
Consequently, the REA must concentrate its perturbations on safety-critical moments.
We then formulate the REA’s decision-making as the following optimization problem
\begin{align}
    \max_{\pi_{\mathrm{adv}}} \quad & \mathbb{E}_{\tau \sim \pi_{\mathrm{adv}}} \left[ \sum_{t=0}^{T-1} \gamma^{t} \, r^{\mathrm{adv}} \left(s_t^{\mathrm{adv}}, a_t^{\mathrm{adv}}\right) \right] , \label{eq:adv_objective} \\[2pt]
    \text{s.t.} \quad 
    & r^{\mathrm{adv}}_t = c(\tilde{s}^{\mathrm{def}}_t, \tilde{a}^{\mathrm{def}}_t), 
    \tag{\ref{eq:adv_objective}a} \label{eq:r^adv} \\ 
    & c(\tilde{s}^{\mathrm{def}}_t, \tilde{a}^{\mathrm{def}}_t) = \mathbb{I}\!\left(\text{safety-critical failure occurs at } t\right),
\tag{\ref{eq:adv_objective}b}
\label{eq:cost} \\
    & a_t^{\mathrm{adv}} = (x_t, u_t^{\mathrm{adv}}) \sim \pi_{\mathrm{adv}}(s_t^{\mathrm{adv}}),
\tag{\ref{eq:adv_objective}c}
\label{eq:a_t_adv} \\
    & \delta_t = AG(\pi_\mathrm{def}, s^{\mathrm{def}}_t, u_t^\mathrm{adv}), 
    \tag{\ref{eq:adv_objective}d} \label{eq:delta} \\ 
    & \tilde{s}^{\mathrm{def}}_t = s^{\mathrm{def}}_t + x_t\delta_t, 
    \tag{\ref{eq:adv_objective}e} \label{eq:s'} \\ 
    & \tilde{a}^{\mathrm{def}}_t = \pi_{\mathrm{def}}(\tilde{s}^{\mathrm{def}}_t), 
    \tag{\ref{eq:adv_objective}f} \label{eq:a'} \\ 
    & \sum_{t=0}^{T-1} x_t \le N_{\mathrm{budget}}.
    \tag{\ref{eq:adv_objective}g} \label{eq:const_attack_gen}
\end{align}
The REA aims to learn an optimal policy $\pi_{\mathrm{adv}}^{\star}$ that selects attack actions to maximize the cumulative adversarial reward, defined by a binary cost function $c(\cdot)$, as in~\eqref{eq:adv_objective} and~\eqref{eq:r^adv}.
Specifically, the cost $c(\tilde{s}^{\mathrm{def}}_t, \tilde{a}^{\mathrm{def}}_t)=1$ if a safety-critical failure (e.g., a collision) occurs at time step $t$, and $c(\tilde{s}^{\mathrm{def}}_t, \tilde{a}^{\mathrm{def}}_t)=0$ otherwise, as defined in~\eqref{eq:cost}.

At each timestep, the REA selects a composite action $a_t^{\mathrm{adv}} = (x_t, u_t^{\mathrm{adv}})$ based on its observation $s_t^{\mathrm{adv}}$, as defined in~\eqref{eq:a_t_adv}.
$s_t^{\mathrm{adv}} = (s_t^{\mathrm{def}}, (N_{\mathrm{remain},t}, a_t^{\mathrm{def}}))$ comprises the clean environment state $s_t^{\mathrm{def}}$ (Section~\ref{sec:agent_3}) and attack-related context, including the remaining attack budget $N_{\mathrm{remain},t}$ and the RTRA’s clean action $a_t^{\mathrm{def}} = \pi_{\mathrm{def}}(s_t^{\mathrm{def}})$, where $\pi_{\mathrm{def}}$ is the RTRA’s driving policy \cite{fan2025less}.
The binary variable $x_t\in\{0,1\}$ indicates whether an attack is activated at timestep $t$. 
When an attack is activated, the continuous component $u_t^{\mathrm{adv}} \in [-1,1]$ specifies the normalized acceleration control signal that the REA intends the RTRA to execute.

Given $u_t^{\mathrm{adv}}$, the perturbation $\delta_t$ is generated by the adversary-guided perturbation generation function $AG(\cdot)$, subject to~\eqref{eq:delta}.
The perturbed observation $\tilde{s}_t^{\mathrm{def}}$ and the corresponding RTRA action $\tilde{a}_t^{\mathrm{def}}$ are then obtained according to~\eqref{eq:s'} and ~\eqref{eq:a'} to compute $c(\cdot)$.
The REA is further constrained by a finite attack budget $N_{\mathrm{budget}}$, which limits the number of attack activations, as defined in~\eqref{eq:const_attack_gen} and forces perturbations to be sparse and focused on safety-critical moments.

\subsection{Optimization Problem Formulation of RTRA}
\label{sec:agent_3}
The primary goal of the RTRA is to learn a driving policy that achieves high driving efficiency while maintaining robustness against potential adversarial perturbations.
This objective is formalized by the following optimization problem
\begin{align}
    \max_{\pi_{\mathrm{def}}} \quad & \mathbb{E}_{\tau \sim \pi_{\mathrm{def}}} \left[ \sum_{t=0}^{T-1} \gamma^{t} \, r^{\mathrm{def}} \left(\tilde{s}^{\mathrm{def}}_t, \tilde{a}^{\mathrm{def}}_t\right) \right] , \label{eq:def_objective} \\[2pt]
    \text{s.t.} \quad
    & \tilde{s}^\mathrm{def}_t = s^\mathrm{def}_t + x_t\delta_t,
    \tag{\ref{eq:def_objective}a} \label{eq:def_s_tilde} \\
    & \tilde{a}^\mathrm{def}_t = \pi_{\mathrm{def}}(\tilde{s}^\mathrm{def}_t),
    \tag{\ref{eq:def_objective}b} \label{eq:def_a} \\
    & \mathbb{E}
    \Big[
    D_{\mathrm{KL}}
    \big(
    \pi_{\mathrm{def}}(\cdot \mid s_t^{\mathrm{def}})
    \;\|\;
    \pi_{\mathrm{def}}(\cdot \mid \tilde{s}_t^{\mathrm{def}})
    \big)
    \;\Big|\;
    x_t = 1
    \Big]
    \le \epsilon_{\mathrm{def}}.
    \tag{\ref{eq:def_objective}c} \label{eq:def_constraint_val}
\end{align}
In \eqref{eq:def_objective}, the RTRA seeks to learn a policy $\pi_{\mathrm{def}}$ that maximizes the expected cumulative driving reward.
Regarding reward design, the RTRA’s reward is designed to balance driving efficiency and safety, which is defined as
\begin{equation}
r^{\mathrm{def}}(\tilde{s}^\mathrm{def}_t, \tilde{a}^\mathrm{def}_t)
=
\frac{v_t}{v_{\max}}
-
c(\tilde{s}^\mathrm{def}_t, \tilde{a}^\mathrm{def}_t),
\label{eq:def_reward}
\end{equation}
where $v_t$ is the longitudinal speed of the ego vehicle, $v_{\max}$ denotes the reference maximum speed used for normalization.

The observation $s^\mathrm{def}_t$ of the RTRA aggregates kinematic information from both the ego vehicle and its surrounding traffic. 
Specifically, it encodes the ego vehicle's speed and heading, alongside the relative distance, orientation, speed, and velocity direction of the six nearest neighboring vehicles within a 200-meter range across six directions (front, rear, front-left, rear-left, front-right, and rear-right).
These components collectively form a 26-dimensional feature vector serving as the basis for the observation $\tilde{s}^\mathrm{def}_t$, upon which the RTRA determines the action $a_t^{\mathrm{def}}$, as defined in~\eqref{eq:def_s_tilde} and~\eqref{eq:def_a}.
The action $a_t^{\mathrm{def}} \in [-1,1]$ represents the normalized longitudinal acceleration of the ego vehicle and shares the same control semantics as the adversarial target action $u_t^{\mathrm{adv}}$.

Constraint~\eqref{eq:def_constraint_val} enforces a consistency constraint on the RTRA’s policy by limiting the deviation between its responses to the clean observation $s^{\mathrm{def}}_t$ and the perturbed observation $\tilde{s}^{\mathrm{def}}_t$.
By preventing adversarial perturbations from causing abrupt changes in the policy, this constraint stabilizes policy learning and avoids overly sensitive reactions.
The constraint is applied only to adversarial samples ($x_t = 1$), as benign samples induce zero deviation and would otherwise dilute the effect of the constraint.

\section{Methodology}
\label{sec:method}
This section presents the concrete implementation of the REA and the RTRA in Sections~\ref{sec:adversary} and~\ref{sec:agent}, respectively. The overall CARRL framework is then detailed in Section~\ref{sec:lfrarl_overall}.

\subsection{REA Design}
\label{sec:adversary}
The REA is trained using proximal policy optimization (PPO)~\cite{schulman2017proximal}. 
PPO is well suited for adversarial policy learning, as its trust-region clipping stabilizes training by preventing abrupt policy updates, and it naturally supports mixed discrete–continuous action spaces.
Notably, the REA is algorithm-agnostic within the class of on-policy methods that support hybrid action spaces, allowing alternative algorithms to be integrated without structural changes.

The REA adopts an actor–critic architecture consistent with standard PPO. 
The actor network $\pi_{\mathrm{adv}}(\theta_{\mathrm{adv}})$ is a policy network that maps the REA’s observation to actions.
In parallel, a critic network $V(\phi_{\mathrm{adv}})$ estimates the state value function to compute the generalized advantage estimation (GAE), which serves to reduce the variance of policy gradient updates.
Under the PPO framework, the REA policy $\pi_{\mathrm{adv}}$ is updated by maximizing a clipped surrogate objective
\begin{equation}
J_{\mathrm{actor}}(\theta_{\mathrm{adv}})
= \mathbb{E} \left[
\min\!\left(
\rho_t(\theta_{\mathrm{adv}})\hat{A}_t,
\rho^{\mathrm{clip}}_t(\theta_{\mathrm{adv}})\hat{A}_t
\right)
\right],
\label{eq:adv_actor}
\end{equation}
where $\hat{A}_t$ denotes the advantage estimate at time step $t$, computed using GAE as
\begin{equation}
\begin{aligned}
\hat{A}_{t} 
&= \sum_{l=0}^{T-t-1} (\gamma \lambda)^{l} \, \delta_{t+l}, \\
\delta_{t} 
&= r^{\mathrm{adv}}_{t} 
  + \gamma V(s^{\mathrm{adv}}_{t+1};\phi_{\mathrm{adv}}^{\mathrm{old}}) 
  - V(s^{\mathrm{adv}}_{t};\phi_{\mathrm{adv}}^{\mathrm{old}}).
\end{aligned}
\end{equation}
The probability ratio $\rho_t(\theta_{\mathrm{adv}})$ is defined as
\begin{equation}
    \rho_t(\theta_{\mathrm{adv}}) =
\frac{\pi_{\mathrm{adv}}(a_t^{\mathrm{adv}} \mid s_t^{\mathrm{adv}};\theta_{\mathrm{adv}})}
     {\pi_{\mathrm{adv}}(a_t^{\mathrm{adv}} \mid s_t^{\mathrm{adv}};\theta_{\mathrm{adv}}^{\mathrm{old}})},
\end{equation}
and $\rho^{\mathrm{clip}}_t(\theta_{\mathrm{adv}}) = 
\mathrm{clip}\!\left(\rho_t(\theta_{\mathrm{adv}}), 1-\epsilon, 1+\epsilon\right)$
restricts the policy update within a trust region controlled by the clipping parameter $\epsilon$.

The critic network is trained by minimizing a value regression loss
\begin{equation}
\mathcal{L}_{\mathrm{critic}}(\phi_{\mathrm{adv}}) =
\mathbb{E}\!\left[
\big(V(s^{\mathrm{adv}}_{t};\phi_{\mathrm{adv}}) - \hat{V}_{t}\big)^{2}
\right],
\label{eq:critic}
\end{equation}
where $\hat{V}_{t}$ denotes the estimated return, computed as
\begin{equation}
\hat{V}_t \;=\; \hat{A}_t \;+\; V(s^{\mathrm{adv}}_t;\phi_{\mathrm{adv}}^{old}).
\end{equation} 

In standard PPO, both action components are optimized at every timestep via the joint log-probability. However, since $u_t^{\mathrm{adv}}$ influences the environment only when $x_t = 1$, updating it during non-attack steps introduces spurious gradients unrelated to any causal effect on returns.
To address this issue, we adopt a decoupled optimization mechanism~\cite{mo2022attacking} that decomposes the adversarial policy $\pi_{\mathrm{adv}}$ into two sub-policies corresponding to the attack trigger ($\pi_x$) and the target action ($\pi_u$). The sub-policies are implemented as an attack-trigger head and a target-action head. Each head is optimized using its own probability ratio, defined as
\begin{equation}
\begin{aligned}
\rho_t^{x}(\theta_{\mathrm{adv}})
&=
\frac{\pi_{x}(x_t \mid s_t^{\mathrm{adv}};\theta_{\mathrm{adv}})}
     {\pi_{x}(x_t \mid s_t^{\mathrm{adv}};\theta_{\mathrm{adv}}^{\mathrm{old}})}, \\
\rho_t^{u}(\theta_{\mathrm{adv}})
&=
\frac{\pi_{u}(u_t^{\mathrm{adv}} \mid s_t^{\mathrm{adv}};\theta_{\mathrm{adv}})}
     {\pi_{u}(u_t^{\mathrm{adv}} \mid s_t^{\mathrm{adv}};\theta_{\mathrm{adv}}^{\mathrm{old}})}.
\end{aligned}
\end{equation}
Since the target action $u_t^{\mathrm{adv}}$ influences the environment only when an attack is active, its update must be conditioned on the trigger decision to avoid spurious gradients.
Accordingly, the actor objective is reformulated as
\begin{equation}
J^{\mathrm{def}}_{\text{actor}}(\theta_{\mathrm{adv}})
=
J_{\text{actor}}^{x}(\theta_{\mathrm{adv}})
+
x_tJ_{\text{actor}}^{u}(\theta_{\mathrm{adv}}),
\label{eq:adv_final_actor}
\end{equation}
where $J_{\text{actor}}^{x}$ and $J_{\text{actor}}^{u}$ denote the standard PPO clipped objectives for the trigger and target-action heads, respectively. 
This decoupled formulation ensures that gradients for the target action head are computed only when it has a causal effect on the environment, enabling stable and causally consistent learning of both attack timing and attack content.

\begin{algorithm}[!htbp]
\SetAlgoLined
\KwIn{Clean observation $s^{\mathrm{def}}_t$; target action $u^{\mathrm{adv}}_t$; the RTRA policy $\pi_{\mathrm{def}}$; step size $\alpha_\delta$; bound $\epsilon_{AG}$}
\KwOut{Perturbation $\delta_t$}
Initialize $\delta_t \leftarrow 0$

\For{each iteration}
{
    $\mathcal{L}_{\mathrm{AG}}(\delta_t) \leftarrow \big\| u^{\mathrm{adv}}_t - \pi_{\mathrm{def}}(s^{\mathrm{def}}_t + \delta_t) \big\|^2$;

    $\delta_t \leftarrow \mathrm{clip}\big(\delta_t - \alpha_\delta \cdot \mathrm{sign}(\nabla_\delta \mathcal{L}_{\mathrm{AG}}(\delta_t)),\, -\epsilon_{AG},\, +\epsilon_{AG}\big)$;
}
\caption{Adversary-Guided Perturbation Generation $AG(\cdot)$}
\label{alg:ag}
\end{algorithm}

When an attack is activated ($x_t=1$), the REA generates a perturbation through $AG(\cdot)$. Given the target action $u_t^{\mathrm{adv}}$, $AG(\cdot)$ produces $\delta_t$ by optimizing the following objective
\begin{equation}
\mathcal{L}_{\mathrm{AG}}(\delta_t) = \big\|u^{\mathrm{adv}}_t - \pi_{\mathrm{def}}(s^{\mathrm{def}}_t + \delta_t)\big\|^2.
\end{equation}
Here, the objective is to minimize the squared $L_2$ distance between the target action and the victim policy's output under the perturbed state.
We optimize $\delta$ via the basic iterative method (BIM)~\cite{kurakin2018adversarial}. At each iteration, $\delta_t$ is updated as
\begin{equation}
\delta_t \leftarrow \mathrm{clip}\big(\delta_t - \alpha_\delta \cdot \mathrm{sign}(\nabla_\delta \mathcal{L}_{\mathrm{AG}}(\delta_t)),\ -\epsilon_{AG},\ +\epsilon_{AG}\big),
\end{equation}
where $\alpha_\delta$ is the step size and $\mathrm{clip}(\cdot)$ constrains the perturbation within the $L_\infty$-bound of $\epsilon_{\mathrm{AG}}$. 
Algorithm~\ref{alg:ag} summarizes this procedure.

\subsection{RTRA Design}
\label{sec:agent}
The RTRA is implemented using soft actor-critic (SAC)~\cite{haarnoja2018soft}.
SAC is adopted for its off-policy learning capability and entropy regularization, enabling efficient use of sparse adversarial transitions and stable policy behavior.
The design remains compatible with other off-policy continuous-control algorithms.

\subsubsection{Soft Actor-Critic Backbone}
We first detail the network architecture and update procedure of the standard SAC. The RTRA employs an Actor-Critic structure, comprising a stochastic policy network $\pi_{\mathrm{def}}$ parameterized by $\theta_{\mathrm{def}}$, and a pair of critic networks $\{Q_{\mathrm{def},1}, Q_{\mathrm{def},2}\}$ parameterized by $\phi_{\mathrm{def}} = \{\phi_{\mathrm{def},1}, \phi_{\mathrm{def},2}\}$ to mitigate value overestimation. Additionally, target critic networks $\{Q'_{\mathrm{def},1}, Q'_{\mathrm{def},2}\}$ are maintained for training stability.

At each update step, the critic networks are trained by minimizing the entropy-regularized Bellman residual. Given a transition $\tau=(\tilde{s}^{\mathrm{def}}_t, s^{\mathrm{def}}_t, a_t^{\mathrm{def}}, r_t^{\mathrm{def}}, s^{\mathrm{def}}_{t+1})$ sampled from the replay buffer $\mathcal{D}_{\mathrm{def}}$, the target value is computed as
\begin{equation} 
\label{eq:critic_target} 
\begin{aligned} 
y_t^{\mathrm{def}} = r_t^{\mathrm{def}} + \gamma \bigg( 
&\min_{j=1,2} Q'_{\mathrm{def},j}(s^{\mathrm{def}}_{t+1}, a^{\mathrm{def}}_{t+1}; \phi'_{\mathrm{def},j}) \\ 
&- \alpha \log \pi_{\mathrm{def}}(a^{\mathrm{def}}_{t+1} \mid s^{\mathrm{def}}_{t+1}; \theta_{\mathrm{def}}) \bigg).
\end{aligned} 
\end{equation} 
Here, $\phi'_{\mathrm{def},j}$ denotes the parameters of the $j$-th target critic network, $\gamma$ is the discount factor, and $\alpha$ is the temperature parameter. The next action is sampled from the current policy $a^{\mathrm{def}}_{t+1} \sim \pi_{\mathrm{def}}(\cdot \mid s^{\mathrm{def}}_{t+1}; \theta_{\mathrm{def}})$.
The critic loss is then defined as:
\begin{equation}
\mathcal{L}^{\mathrm{def}}_{\text{critic}}(\phi_{\mathrm{def}})
=
\mathbb{E}_{\tau \sim \mathcal{D}_{\mathrm{def}}}
\Bigg[
\sum_{j=1}^2
\Big(
Q_{\mathrm{def},j}(\tilde{s}^{\mathrm{def}}_t, \tilde{a}_t^{\mathrm{def}}; \phi_{\mathrm{def},j})
-
y_t^{\mathrm{def}}
\Big)^2
\Bigg].
\end{equation}
The parameters $\phi_{\mathrm{def}}$ are updated via gradient descent to minimize $\mathcal{L}^{\mathrm{def}}_{\text{critic}}$.

To maximize the expected return and entropy, the actor is updated by minimizing the loss function defined as
\begin{equation}
\label{eq:actor_loss}
\begin{aligned}
\mathcal{L}^{\mathrm{def}}_{\text{actor}}(\theta_{\mathrm{def}})
&=
\mathbb{E}_{\tilde{s}^{\mathrm{def}}_t \sim \mathcal{D}_{\mathrm{def}}}
\Big[
\alpha \log \pi_{\mathrm{def}}(\tilde{a}^{\mathrm{def}}_t \mid \tilde{s}^{\mathrm{def}}_t; \theta_{\mathrm{def}}) \\
&\quad -
\min_{j=1,2}
Q_{\mathrm{def},j}(\tilde{s}^{\mathrm{def}}_t, \tilde{a}^{\mathrm{def}}_t; \phi_{\mathrm{def},j})
\Big].
\end{aligned}
\end{equation}
where $\tilde{a}^{\mathrm{def}}_t \sim \pi_{\mathrm{def}}(\cdot|\tilde{s}^{\mathrm{def}}_t;\theta_{\mathrm{def}})$.

\subsubsection{Dual Replay Buffer}
Although standard SAC is highly sample-efficient, its performance in adversarial settings is limited by the scarcity and imbalance of adversarial data.
Due to the constrained attack budget, adversarial samples are scarce and easily diluted under uniform replay sampling, limiting their contribution to robust policy learning.

To address this issue, we introduce a DRB to explicitly decouple benign and perturbed transitions. 
Transitions are categorized according to the REA’s trigger action $x_t$ and stored in the corresponding buffers as 
\begin{equation}
\tau_t \in
\begin{cases}
\mathcal{D}^{\text{normal}}_{\mathrm{def}}, & \text{if } x_t = 0 \ (\text{benign}), \\[3pt]
\mathcal{D}^{\text{attack}}_{\mathrm{def}}, & \text{if } x_t = 1 \ (\text{perturbed}).
\end{cases}
\end{equation}
During training, a mini-batch $\mathcal{B}$ of size $N_{\text{batch}}$ is constructed by sampling from $\mathcal{D}^{\text{normal}}_{\mathrm{def}}$ and $\mathcal{D}^{\text{attack}}_{\mathrm{def}}$. 
A mixing ratio $\beta \in [0,1]$ controls the proportion of adversarial samples in each batch. 
Specifically,
\begin{equation}
\mathcal{B}
=
\mathcal{B}_{\text{normal}}
\cup
\mathcal{B}_{\text{attack}},
\qquad
\lvert \mathcal{B}_{\text{attack}} \rvert
=
\lfloor \beta N_{\text{batch}} \rfloor,
\label{eq:drb}
\end{equation}
where $\mathcal{B}_{\text{normal}} \subset \mathcal{D}^{\text{normal}}_{\mathrm{def}}$ and $\mathcal{B}_{\text{attack}} \subset \mathcal{D}^{\text{attack}}_{\mathrm{def}}$.
This sampling strategy ensures sufficient exposure to high-risk scenarios while preserving performance in benign traffic.
Leveraging the transitions sampled from the DRB, the critic loss is computed as
\begin{equation}
\label{eq:drb_critic_loss}
\mathcal{L}^{\mathrm{def}}_{\text{critic}}(\phi_{\mathrm{def}})=\mathbb{E}_{\tau \sim \mathcal{B}}
\Bigg[
\sum_{j=1}^2
\Big(
Q_{\mathrm{def},j}(\tilde{s}^{\mathrm{def}}_t, \tilde{a}_t^{\mathrm{def}}; \phi_{\mathrm{def},j})-y_t^{\mathrm{def}}\Big)^2\Bigg].
\end{equation}

\subsubsection{Consistency-Constrained Policy Optimization}
While the DRB ensures that the RTRA is sufficiently exposed to adversarial scenarios, training on these samples alone does not guarantee robustness.
To explicitly enforce robustness against adversarial perturbations, the policy is further required to satisfy a consistency constraint that bounds the deviation between the action distributions induced by a benign observation $s^{\mathrm{def}}_t$ and its perturbed counterpart $\tilde{s}^{\mathrm{def}}_t$. 
This constraint $c_{\text{consist}}$ is formalized as
\begin{equation}
\label{eq:constraint_term}
\mathbb{E}_{\tau \in \mathcal{B}_{\mathrm{attack}}}
\Bigg[
D_{\mathrm{KL}} \Big(
\pi_{\mathrm{def}}(\cdot \mid s_t^{\mathrm{def}}) \;\big\|\;
\pi_{\mathrm{def}}(\cdot \mid \tilde{s}_t^{\mathrm{def}})
\Big)
\Bigg]
\le \epsilon_{\mathrm{def}},
\end{equation}
where $\epsilon_{\mathrm{def}}$ is a predefined tolerance threshold. 
Crucially, we restrict this constraint to the adversarial subset $\mathcal{B}_{\text{attack}}$.
We evaluate this constraint exclusively on $\mathcal{B}_{\text{attack}}$ to avoid dilution by zero-loss benign samples. This decouples the interpretation of $\epsilon_{\mathrm{def}}$ from the replay buffer's mixing ratio $\beta$.

We incorporate this constraint into the actor's objective via Lagrangian relaxation. 
The final loss function balances the standard SAC objective with the consistency requirement, which is defined as
\begin{equation}
\label{eq:actor_total_loss}
\mathcal{L}^{\mathrm{def}}_{\text{actor-lag}}(\theta_{\mathrm{def}})
=
\mathcal{L}^{\mathrm{def}}_{\text{actor}}(\theta_{\mathrm{def}})
+
\lambda_{\mathrm{def}}\left(c_{\text{consist}}(\theta_{\mathrm{def}})-\epsilon_{\mathrm{def}}\right),
\end{equation}
where $\lambda_{\mathrm{def}}$ is the dual variable updated via dual gradient ascent
\begin{equation}
\lambda_{\mathrm{def}} \leftarrow \max\Big( 0,  \lambda_{\mathrm{def}} + \alpha_{\lambda} \big( c_{\text{consist}}(\theta_{\mathrm{def}}) - \epsilon_{\mathrm{def}} \big) \Big),
\label{eq:lambda}
\end{equation}
where $\alpha_\lambda$ denotes the learning rate.
This design ensures that the regularization term is triggered primarily by adversarial samples, thereby tightly regularizing the policy in high-risk, perturbed scenarios while preserving its optimality in clean scenarios.

\subsection{Overall Approach}
\label{sec:lfrarl_overall}
We now present the complete training procedure of CARRL. 
To promote the co-evolution of the REA and the RTRA, we adopt an iterative training paradigm~\cite{pinto2017robust}.
The training proceeds over $N_{\text{iter}}$ iterations, alternating between two distinct phases: agent training and adversary training.
This alternating training process continuously exposes the RTRA to evolving adversarial behaviors, avoiding overfitting to fixed attack patterns and enabling convergence toward a robust policy $\pi_{\mathrm{def}}^{\star}$. 
\section{Experiments}
\label{sec:experiments}
This section evaluates the performance and robustness of CARRL in an unprotected left-turn intersection scenario implemented in the SUMO traffic simulator~\cite{lopezMicroscopicTrafficSimulation2018}. This scenario represents a challenging and safety-critical driving task, making it well suited for assessing robustness under adversarial perturbations. 

We first describe the experimental environment and task setup, followed by the baseline methods, evaluation metrics, and training details in Sections~\ref{sec:setting}-~\ref{sec:training}, respectively. These descriptions provide the necessary context for fair comparison and ensure the reproducibility of our results. 
We organize our empirical evaluation to systematically address the following research questions:
\begin{itemize}
    \item \textbf{RQ1}: How does CARRL perform compared to classical and state-of-the-art baseline methods?
    \item \textbf{RQ2}: How do unrestricted, continuous adversarial perturbations affect the robustness and behavior of the learned policy?
    \item \textbf{RQ3}: How well does the driving policy learned by CARRL generalize to environments with conditions different from the training setting?
\end{itemize}

\begin{table}[!htbp]
    \caption{Summary of Experimental Parameters}
    \label{tab:exp_params}
    \centering
    \begin{tabular}{lll}
        \toprule
        Category & Parameter & Value \\
        \midrule
        \multirow{4}{*}{Env} 
            & Max speed $v_{\max}$                    & $15~\text{m/s}$ \\
            & Acceleration range                      & $[-7.6, 7.6]~\text{m/s}^2$ \\
            & Traffic density                   & $0.3,0.5,0.7$ \\
            & Max steps per episode $T$  & $30$ \\
        \midrule
        \multirow{10}{*}{DRL}
            & Learning rate                    & $1 \times 10^{-4}$ \\
            & Temperature parameter $\alpha$          & $0.1$ \\
            & Lagrange multiplier stepsize $\alpha_{\lambda}$        & $5 \times 10^{-5}$ \\
            & Discount factor $\gamma$                & $0.99$ \\
            & Clipping parameter $\epsilon$ & $0.2$ \\
            & Constraint threshold $\epsilon_{\text{def}}$ & $0.1$ \\
            & Batch size                              & $64$ \\
            & Replay buffer size                      & $1 \times 10^{6}$ \\
            & Adversarial sample ratio $\beta$  & $0.5$ \\
        \midrule
        \multirow{4}{*}{AG}
            & Perturbation magnitude $\epsilon_{\text{AG}}$       & $0.03, 0.05, 0.07$ \\
             & Max attack steps $N_{\text{attack}}$    & $5$ \\
            & Number of iterations                    & $50$ \\
            & Update step size $\alpha_{\delta}$      & $\epsilon_\mathrm{AG} / 50$ \\
        \bottomrule
    \end{tabular}
\end{table}

\subsection{Environment Setting}
\label{sec:setting}
The lane-changing and longitudinal speed control behaviors of surrounding vehicles are governed by the built-in intelligent driver model (IDM)~\cite{albeaik2022limitations}. 
Each vehicle is configured with a maximum speed of $15\,\mathrm{m/s}$, and its longitudinal acceleration is constrained within the range $[-7.6,\,7.6]\,\mathrm{m/s^2}$~\cite{heTrustworthyAutonomousDriving2024a}. 
Unless otherwise specified, the per-second arrival probability of vehicles, i.e., the traffic density, is fixed at 0.5.

\subsection{Baselines}
\label{sec:baselines}
To ensure a comprehensive and fair evaluation, we benchmark the proposed approach against representative state-of-the-art DRL-based AD methods across three categories. First, we include widely adopted \emph{Vanilla RL} methods, namely PPO~\cite{schulman2017proximal}, SAC~\cite{haarnoja2018soft}, and TD3~\cite{fujimoto2018addressing}, which cover both on-policy and off-policy learning paradigms and serve as strong baselines for nominal driving performance. Second, we compare against \emph{Safe RL} methods that explicitly incorporate safety constraints into policy optimization, including SAC\_Lag~\cite{ha2021learning} and the state-of-the-art method, FNI~\cite{He2023FearNeuroInspiredRL}. Finally, we evaluate CARRL against state-of-the-art \emph{Robust RL} methods, represented by DARRL~\cite{heTrustworthyAutonomousDriving2024a}, which integrates continuous adversarial perturbations to enhance decision stability in challenging driving scenarios.

\subsection{Metrics}
\label{sec:metrics}
We evaluate the agent’s performance over $N=500$ independent episodes per training seed using four complementary metrics. The success rate (SR) measures the proportion of episodes in which the ego vehicle reaches the target lane without collision, serving as the primary indicator of task completion. The collision rate (CR) quantifies the frequency of safety violations, defined as the proportion of episodes involving a collision. The driving efficiency (DE) is measured by the average speed of the ego vehicle over an episode, reflecting the agent’s ability to navigate efficiently. In addition to these performance metrics, we assess algorithmic stability (AS) by examining the variance of evaluation metrics across independent training seeds. In safety-critical AD, this analysis is essential to verify that the observed robustness is consistent and reproducible, rather than arising from favorable initialization or stochastic training effects.

\begin{table*}[!htbp]
  \centering
  \caption{Statistical Results Under Different Perturbation Magnitudes $\epsilon_\mathrm{AG}$}
  \label{tab:attack_results}
  \small
  \renewcommand{\arraystretch}{1.2}
  \setlength{\tabcolsep}{4pt}
  \begin{tabular}{l l ccc|cc|c|c}
    \toprule
    Condition & Metric &
    PPO & SAC & TD3 & SAC\_Lag & FNI & DARRL & \textbf{CARRL (Ours)} \\
    \midrule
    \multirow{3}{*}{w.o attacks}
      & SR &
         \textbf{\textcolor{red}{100.00$\pm$0.00}} &
        89.17$\pm$9.00 &
        87.33$\pm$10.98 &
        96.33$\pm$6.35 &
         \textbf{\textcolor{red}{100.00$\pm$0.00}} &
        81.67$\pm$7.69 &
        \textbf{\textcolor{red}{100.00$\pm$0.00}} \\
      & CR &
         \textbf{\textcolor{red}{0.00$\pm$0.00}} &
        10.83$\pm$9.00 &
        12.67$\pm$10.98 &
        3.67$\pm$6.35 &
         \textbf{\textcolor{red}{0.00$\pm$0.00}} &
        18.33$\pm$7.69 &
        \textbf{\textcolor{red}{0.00$\pm$0.00}} \\
      & DE &
        \textcolor{red}{12.97$\pm$0.01} &
        12.81$\pm$0.05 &
        12.76$\pm$0.21 &
        12.68$\pm$0.30 &
        12.62$\pm$0.01 &
        12.73$\pm$0.22 &
        10.98$\pm$0.01 \\
    \midrule
    \multirow{3}{*}{$\epsilon_{\text{AG}} = 0.03$}
      & SR &
        17.83$\pm$0.29 &
        43.00$\pm$3.50 &
        42.33$\pm$23.86 &
        68.00$\pm$10.33 &
        43.67$\pm$17.48 &
        48.17$\pm$15.04 &
        \textbf{\textcolor{red}{96.33$\pm$0.58}} \\
      & CR &
        82.17$\pm$0.29 &
        56.67$\pm$3.75 &
        57.67$\pm$23.86 &
        31.83$\pm$10.37 &
        56.33$\pm$17.48 &
        51.83$\pm$15.04 &
        \textbf{\textcolor{red}{3.67$\pm$0.58}} \\
      & DE &
         \textbf{\textcolor{red}{13.40$\pm$0.00}} &
        13.03$\pm$0.00 &
        12.98$\pm$0.64 &
        12.33$\pm$0.44 &
        13.13$\pm$0.06 &
        12.40$\pm$0.69 &
        10.85$\pm$0.02 \\
    \midrule
    \multirow{3}{*}{$\epsilon_{\text{AG}} = 0.05$}
      & SR &
        12.67$\pm$0.58 &
        12.83$\pm$2.36 &
        17.50$\pm$13.00 &
        48.17$\pm$5.03 &
        23.33$\pm$11.85 &
        42.17$\pm$24.01 &
        \textcolor{red}{\textbf{95.60 $\pm$ 0.28}} \\
      & CR &
        87.33$\pm$0.58 &
        87.00$\pm$2.65 &
        82.50$\pm$13.00 &
        50.50$\pm$7.21 &
        76.67$\pm$11.85 &
        57.83$\pm$24.01 &
        \textcolor{red}{\textbf{4.40 $\pm$ 0.28}} \\
      & DE &
         \textbf{\textcolor{red}{13.44$\pm$0.00}} &
        13.17$\pm$0.02 &
        13.04$\pm$0.98 &
        11.72$\pm$0.98 &
        13.29$\pm$0.11 &
        12.14$\pm$1.13 &
        10.91$\pm$0.01 \\
    \midrule
    \multirow{3}{*}{$\epsilon_{\text{AG}} = 0.07$}
      & SR &
        10.00$\pm$0.00 &
        9.17$\pm$3.33 &
        10.50$\pm$8.79 &
        38.67$\pm$0.29 &
        12.67$\pm$7.65 &
        34.83$\pm$14.29 &
        \textbf{\textcolor{red}{61.33$\pm$0.29}} \\
      & CR &
        90.00$\pm$0.00 &
        90.83$\pm$3.33 &
        89.50$\pm$8.79 &
        61.33$\pm$0.29 &
        87.33$\pm$7.65 &
        65.17$\pm$14.29 &
        \textbf{\textcolor{red}{38.67$\pm$0.29}} \\
      & DE &
         \textbf{\textcolor{red}{13.39$\pm$0.07}} &
        13.32$\pm$0.02 &
        12.79$\pm$1.19 &
        11.42$\pm$1.21 &
        13.29$\pm$0.14 &
        12.00$\pm$1.44 &
        11.28$\pm$0.02 \\
    \bottomrule
  \end{tabular}
\end{table*}

\begin{figure}[!htbp]
    \centering
    \includegraphics[width=\linewidth, trim=5 5 5 5,clip]{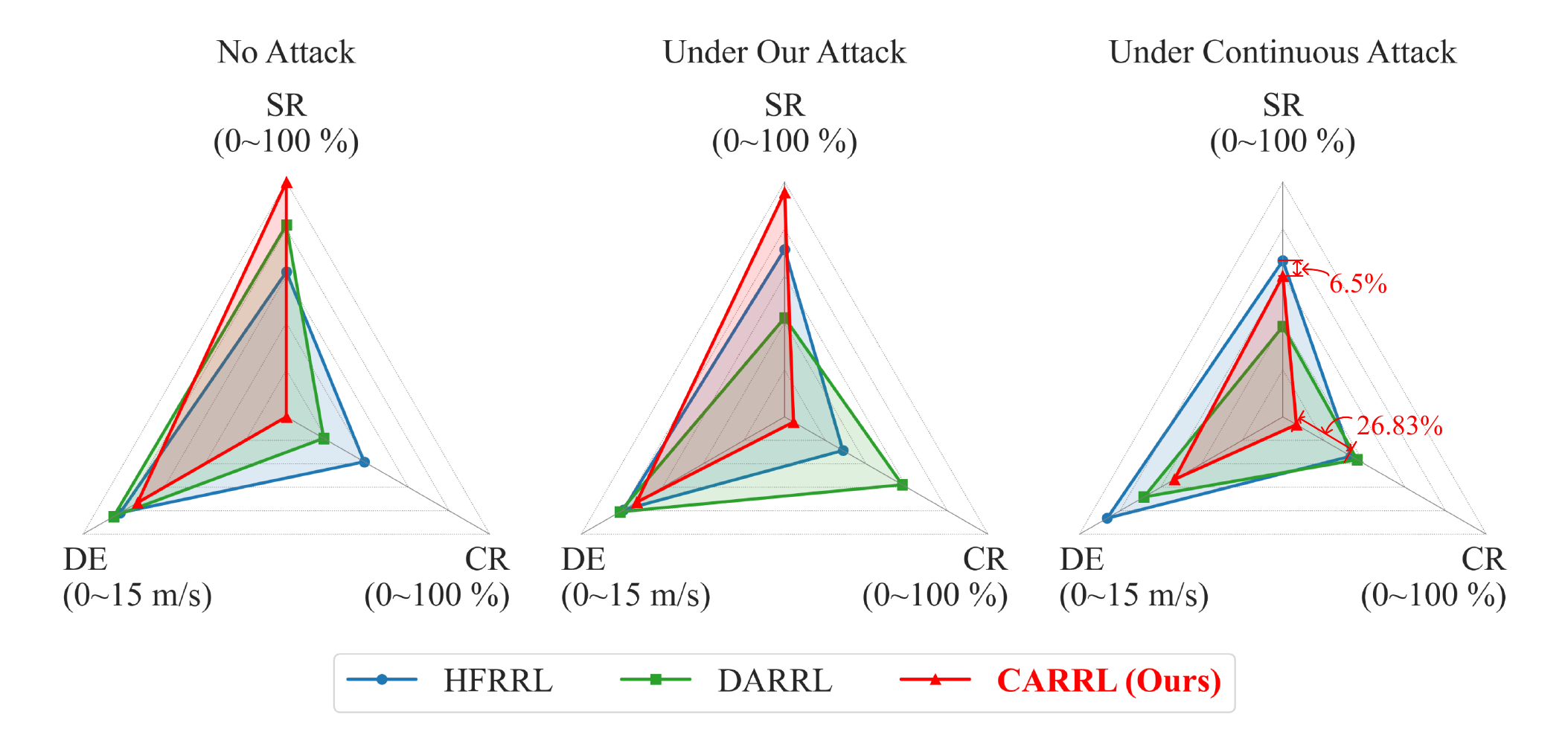}
    \caption{Performance comparison under different attack policies.}
    \label{fig:attack_compare}
\end{figure}

\subsection{Training Details}
\label{sec:training}
All experiments are conducted on a workstation equipped with an Intel Xeon Gold 6230 CPU and an NVIDIA GeForce RTX~4090 GPU. 
Key hyperparameters (summarized in Table~\ref{tab:exp_params}) are selected based on prior literature~\cite{heTrustworthyAutonomousDriving2024a} and carefully tuned to ensure fair and competitive baselines.
Unless otherwise specified, DRL-based agents adopt the default hyperparameter settings from Stable Baselines3~\cite{stable-baselines3}. To account for stochasticity in training, each method is independently trained using five different random seeds.

\subsection{Performance Evaluation (RQ1)}
Table~\ref{tab:attack_results} reports the performance of CARRL and baseline methods under different perturbation magnitudes $\epsilon_{\text{AG}}$, including results in the clean environment. 
Overall, the results demonstrate that CARRL consistently outperforms classical and state-of-the-art baselines in robustness against adversarial perturbations, while maintaining strong performance in the clean environment.

In the clean environment, CARRL, PPO, and FNI all achieve a $100\%$ SR, confirming that CARRL does not degrade the agent's basic driving capability. 
Although CARRL attains a lower DE than PPO, this reflects a trade-off toward more cautious and stable driving behavior, which is well justified in safety-critical scenarios.

When adversarial perturbations are introduced, CARRL shows a substantial advantage over all baselines. 
At $\epsilon = 0.03$, CARRL outperforms the strongest baseline by an average margin of $28.33\%$ in success rate. This advantage becomes even more pronounced at $\epsilon = 0.05$, where the performance gap further widens to over $47.43\%$.
Under the most challenging setting ($\epsilon_{\text{AG}}=0.07$), CARRL remains markedly superior, achieving a SR of 61.33\%.
This is approximately 1.6 times higher than the strongest baseline, SAC\_Lag, which achieves 38.67\%, while vanilla RL methods degrade to near-random performance.
Regarding AS, CARRL outperforms all baselines by consistently maintaining variances in both SR and CR, keeping them below 0.6 across all cases.
This confirms that its robustness improvements are systematic rather than incidental.

\begin{figure*}[!htbp]
    \centering
    \includegraphics[width=\textwidth]{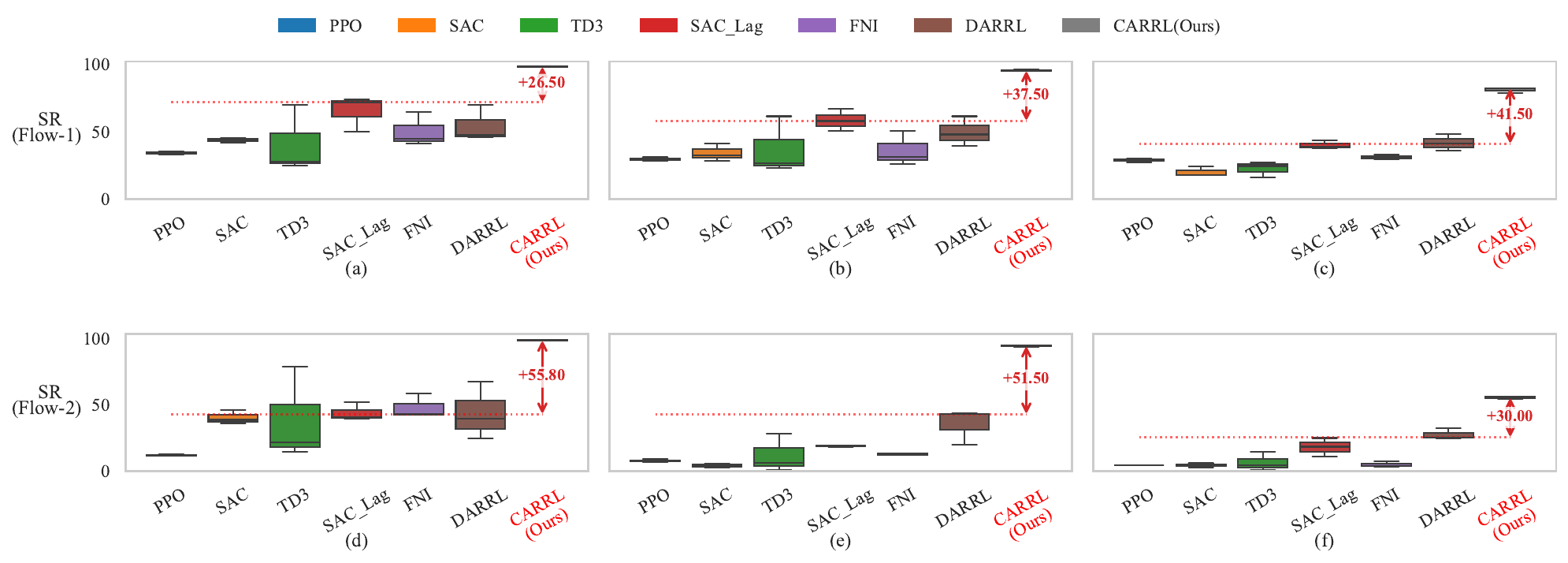}
    \caption{Success rates of different methods under adversarial attacks across varying traffic densities and perturbation magnitudes. The top row corresponds to low-density traffic (Flow-1, $\rho = 0.3$), while the bottom row represents high-density traffic (Flow-2, $\rho = 0.7$). Each column corresponds to a different perturbation magnitude, with $\epsilon_{\text{AG}} = 0.03$, $0.05$, and $0.07$ from left to right.}

    \label{fig:sr_boxplot_density_eps}
\end{figure*}
\subsection{Performance under Continuous Perturbations (RQ2)}
\label{sec:rq2}
To further evaluate the generalization robustness of CARRL under continuous attacks, we introduce a continuous attack scenario for comparison. 
The continuous attack is a variant of our REA with the attack budget constraint removed, allowing perturbations to be applied at every timestep. 
Based on this setup, we construct a new baseline method, high-frequency robust reinforcement learning (HFRRL), a variant of CARRL in which the REA is replaced by the continuous attack adversary during training.

Fig.~\ref{fig:attack_compare} illustrates the performance comparison between CARRL, HFRRL, and DARRL across different attack scenarios. 
We include DARRL as it also serves as a defense paradigm under continuous attacks.
Under both the clean environment and the criticality-aware attack (denoted as 'Our Attack'), CARRL demonstrates a significant performance advantage over HFRRL and DARRL in terms of both SR and CR.
The results highlight the inherent instability of DARRL and HFRRL. 
Specifically, these baselines exhibit overly aggressive driving behaviors across all evaluated scenarios, leading to reckless maneuvers and elevated CRs.
A fundamental cause is the scarcity of clean samples during training. Without sufficient exposure to clean driving dynamics, the agents fail to establish a stable behavioral anchor, causing the learned policies to overfit to adversarial perturbations.
In the continuous attack scenario, CARRL and the baselines do not exhibit a significant divergence in SR. 
Notably, it achieves a substantial 26.83\% reduction in collision rate compared to HFRRL, with only a slight 6.5\% drop in SR.
The consistent robustness of CARRL demonstrates its capability to learn a generalized robust AD policy, establishing its viability for trustworthy operation in complex and unpredictable environments.

\subsection{Generalization Evaluation (RQ3)}
\label{sec:rq3}
We evaluate all methods across varying traffic densities to validate the generalization capability of CARRL to the dynamic nature of real-world driving environments, ensuring robustness amidst shifting traffic conditions.

Across all tested scenarios, CARRL consistently outperforms every baseline method, with the performance gap widening as the perturbation magnitude $\epsilon_{\text{AG}}$ increases. 
In the low-density setting (Flow-1), competitive robust methods such as DARRL and SAC\_Lag retain moderate performance at $\epsilon_{\text{AG}}=0.03$ but degrade sharply as $\epsilon_{\text{AG}}$ increases to $0.07$. 
In contrast, CARRL maintains strong robustness, surpassing the best baseline by margins of $26.50\%$ to $41.50\%$, demonstrating superior robustness to variations in both traffic density and perturbation magnitude.
This advantage becomes even more pronounced under high-density traffic (Flow-2), where most baseline methods suffer near-complete failure. 
Conversely, CARRL maintains a success rate approaching $100\%$ under moderate conditions ($\epsilon_{\text{AG}}=0.03$).
Crucially, even in the most severe scenario ($\epsilon_{\text{AG}}=0.07$), it sustains a performance exceeding $50\%$, proving its ability to preserve functional safety when baseline methods collapse.
Furthermore, its consistently low variance across random seeds highlights superior AS compared to the highly volatile performance of other baseline methods.
These results indicate that CARRL generalizes effectively to unseen, dense traffic conditions while preserving robustness against varying attack strengths, highlighting its practical value for reliable and safe autonomous driving in real-world conditions.
\FloatBarrier
\section{Conclusion}
\label{sec:conclusion}
In this work, we presented CARRL, a novel DRL-based approach for robust AD under sparse yet safety-critical risks. 
CARRL comprises two components (namely, REA and RTRA), and explicitly decouples their reward structures to enable asymmetric learning.
This design allows the REA to pursue worst-case-oriented behaviors by exposing safety-critical failures, while the RTRA learns to balance safety and driving efficiency.
Through selectively identifying and exploiting safety-critical moments, the REA realizes criticality-aware attacks.
To counter the resulting scarcity of adversarial data, the RTRA incorporates the DRB to rebalance the experience distribution and employs the CCPO to anchor behavioral stability under perturbations.
Extensive results show that CARRL consistently outperforms baseline methods in terms of SR, CR, and AS across varying perturbation magnitudes and traffic densities, demonstrating strong robustness.  
Furthermore, its effectiveness under both the proposed REA and continuous attack settings indicates that focused defense at critical moments generalizes to global robustness.
These findings highlight the strong generalization capability of CARRL and its potential to improve the reliability of autonomous driving policies in complex, real-world environments.

\FloatBarrier
\bibliographystyle{jabbrv_IEEEtran}
\bibliography{main}


\end{document}